%% file: dnd-article-template-arxiv.tex
\documentclass[twoside,11pt]{article}

\usepackage{dnd}
\usepackage{times}
\usepackage{multirow}

\dndheading{issue(number)}{year}{firstpage--lastpage}{Elena Musi, Debanjan Ghosh, and Smaranda Muresan}{10.5087/dad.DOINUMBER}

\ShortHeadings{Do concessions increase persuasion}{Musi, Ghosh and Muresan}
\firstpageno{1}

\begin{document}

\title{ChangeMyView Through Concessions: Do Concessions Increase Persuasion?}

\author{\name Elena Musi \email em3202@columbia.edu \\
       \addr Data Science Institute\\
       Columbia University
       \AND
       \name Debanjan Ghosh \email debanjan.ghosh@rutgers.edu \\
       \addr School of Communication and Information\\
       Rutgers University
     \AND 
     \name Smaranda Muresan  \email smara@columbia.edu\\
       \addr Data Science Institute\\
       Columbia University}

\editor{Maite Taboada}
\submitted{07/2017}{04/2018}{MM/YYYY}

\maketitle

\begin{abstract}%
In discourse studies concessions are considered among those argumentative strategies that increase persuasion. We aim to empirically test this hypothesis by calculating the distribution of argumentative concessions in persuasive vs. non-persuasive comments
%Delta-awarded vs. no-Delta discussion threads 
%SM here people might not know directly what is a delta vs non-delta awarded comment
from the ChangeMyView subreddit. This constitutes a challenging task since concessions are not always part of an argument. 
%em: do not always bear an argumentative role and are expressed through polysemous lexical markers.
%To this end, 
Drawing from a theoretically-informed typology of concessions, we 
%first 
conduct an annotation task to label a set of polysemous lexical markers as introducing an argumentative concession or not and we observe their distribution in threads that achieved and did not achieve persuasion. For the annotation, we used both expert and novice annotators. 
%Second, 
With the ultimate goal of conducting the study on large datasets, we present a self-training method to automatically identify argumentative concessions using linguistically motivated features.
%EM after_review: 
We achieve a moderate F1 of 57.4\% on the development set and  46.0\% on the test set via the self-training method. These results are comparable to state of the art results on similar tasks of identifying explicit discourse connective \emph{types} from the Penn Discourse Treebank. 
%EM_after_review
%The analysis of the misclassified occurrences shows that pragmatic knowledge is sometimes needed to grasp the argumentative role played by concessions. 
%While we achieve a moderate F1 of 57.4\% via the self-training method, 
%our subsequent error analysis highlights that the self training method is able to generalize and identify other types of concessions that are argumentative, but were not considered in the annotation guidelines.  
%SM I added that we also did manually labeling of data
% classification experiment to automatically identify argumentative concessions. 
%The training and test data are collected from the ChangeMyView subreddit. 
%As features we have considered those linguistic patterns used in state of the art studies about arguments identifications. Beside those, we have added a lexicon of lexical patterns associated to argumentative concessions, developed through bootstrapping. 
%EM features are inspired from linguistic patterns as well as discourse and argumentation studies. 
%we qualitatively analyze the reasons underlying the difficulties encountered in the classification experiment. 
Our findings from the manual labeling and the classification experiments indicate that the type of argumentative concessions we investigated is almost equally likely to be used in winning and losing arguments from the ChangeMyView dataset. While this result seems to contradict theoretical assumptions, we provide some reasons for this discrepancy related to the ChangeMyView subreddit.

%EM however we also discuss the difficulty of classifying argumentative concessions against other comparative linguistic constructs.  

%collect a corpus using crowdsourcing; we provide a bootstrapping method for building a lexicon of linguistic constructions marking argumentative concessions. We embed the bootstrapped lexicon as features in a classification experiment next to other semantic and pragmatic properties, obtaining 57 accuracy. 
%\em: not sure about what follows
%Our findings indicate that this particular type of argumentative constructions is almost equally likely to be used in winning and losing arguments. While this result seems to contradict theoretical assumptions, we provide some reasons related to ChangeMyView subreddit.
\end{abstract}


\input{introduction}

\input{related}

\input{typology}

\input{empirical}
\input{discussion}
\section{Conclusion and future work}

We tackled the task of empirically validating the theoretically assumed  persuasive role played by specific discourse relations, \emph{concessions}, using the ChangeMyView platform. Drawing from a linguistically-informed typology of concessions, we singled out one type of concessions that prototypically bears an argumentative value. We focused on four discourse markers that constitute 85\% of the overall occurrences of potential markers in our data: \textit{but}, \textit{though}, \textit{however} and \textit{while}. We present a computational model based on self-training using linguistically motivated features.  We achieve a moderate F1 of 57.4\% via the self-training method on the development set and 46.0\% on the test set.  
%EM after_review
%our subsequent error analysis highlights that the self training method is able to generalize and identify other types of concessions that are argumentative, but were not considered in the annotation guidelines used to label our test set. 
Our findings both from the manual labeling (both expert and crowdsourcing annotation) and the system predictions indicate that the type of argumentative concessions we investigate is almost equally likely to be used in winning ($\Delta$-awarded) and losing (no-$\Delta$) arguments.
While this result seems to contradict theoretical assumptions, we provided some reasons related to the ChangeMyView subreddit. This behavior shows that text-genre rules have to be taken into account in the interpretation of rhetorical patterns: When perceived as conventional genre-specific rules, persuasive strategies may happen not to be pragmatically effective. In future work, we plan to validate this explanation by the following three steps: 1) improving the performance of our computational models by  collecting a larger dataset as training data to be able to run the distribution over a much larger dataset; 2) looking at other types of argumentative concessions; and 3) retrieving argumentative concessions in different text-genres. 

\section*{Acknowledgement}
This paper is based on work supported partially by the Advanced Post Doc SNFS Grant for the project “From semantics to Argumentation Mining: a Context-Indipendent Lexicon of Indicators of Argumentative Discourse Relations ” and DARPA-DEFT program. The views expressed are those of the authors and do not reflect the official policy or position of the SNFS, Department of Defense or the U.S. Government. We would like to thank the annotators for their work and the anonymous reviewers for their valuable feedback

%\bibliography{concession}

\end{document}

%% file: introduction.tex
\section{Introduction} \label{section:intro}
A major challenge for \textit{Argument Mining} --- the automatic identification of argumentative structures within discourse--- is the identification of linguistic features which characterize a winning argument. Discourse moves that allow a speaker to achieve persuasion 
have been investigated since the Antiquity, being at the core of \textit{Rhetoric}, and are still a central concern in contemporary discourse studies and \textit{argumentation theory}. The automatic identification of persuasive discourse is also receiving more and more attention in computational linguistics. 
%EM-after review- 
Concessions have been unanimously deemed as strategies which increase persuasion
%Social Science 
%EM_after review 
(Section \ref{section:related}). 
%SM I commented the below and changed to make the link to previous sentence more logical 
%However, an account of the argumentative role played by concessions is so far lacking. 
However, not every concession is part of an argument. We provide a \emph{definition of argumentative concessions} in semantic and pragmatic terms. On this basis, 
%SM There is no link between this statement of not agreed definition and then we empirically test in the next sentence... If there is no definition, what are we testing. Maybe here we say in this paper we provide a typology of concessions that can have an argumentative role and then test empirically whether a certain type of argumentative concessions is persuasive?? 
%This paper aims at unraveling complex linguistic patterns of persuasive arguments starting from theoretically-grounded assumptions. In \textit{Rhetoric},\citep{antaki1999show} \textit{Argumentation Theory} \citep{uzelgun2015managing} and \textit{Discourse Studies} \citep{mann1988rhetorical,dafouz2008pragmatic}, concessions (e.g.,``I agree that this dress is expensive, but I still want to buy it") are unanimously deemed as strategies which achieve persuasion. This paper aims to empirically test this theoretically-informed hypothesis by 
%To empirically test this theoretically-informed hypothesis, e 
we \emph{empirically test the theoretically-informed hypothesis that argumentative concessions work as persuasive strategies} 
%EM-after reviews 
by calculating their distribution 
%of argumentative concessions 
in persuasive vs. non persuasive discourse. An ideal dataset to carry out this empirical investigation is the ChangeMyView subreddit platform, where multiple users negotiate opinions on a certain issue willing to change their point of view through other users' arguments. When their point of view is changed, they award a $\Delta$ point and back it up with a reason. Thus, this platform provides us with a clear user intent --- persuasion --- and a clear signal when a message is perceived as persuasive ($\Delta$ point). 
%EM-after reviews 
Even though it would have been convenient to observe the function of concessions in different datasets, the lack of explicit signals of persuasion prevents us to carry out such a comparison.
%EM: I think from here it is pretty clear why we cannot test this anywhere else. 
%EM---after review 
We use the ChangeMyView dataset released by \citet{tanetal16}.
%$\Delta$-awarded vs. no-$\Delta$ discussion threads in the ChangeMyView corpus introduced by \cite{tanetal16}. 

Our task faces two challenges. First, 
%EM: after reviews
in order for concessions to increase persuasion, they need to be part of an argument. 
%convey an argumentative value. 
However, not every concession 
%presupposes argumentation 
is argumentative \citep{grote1997ma}: The pragmatic function of the sentence ``Although it is December, there is no snow" is not 
%that of 
convincing the hearer about the lack of snow --- which is an observable fact not subject to doubt --- but simply that of expressing surprise for an unusual combination of events. This sentence could work in certain contexts as an argument (e.g., for the standpoint ``Global warming is worse and worse") but does not semantically presuppose any controversy and, thus, the presence of argumentation. 
%SM I removed "reliably" since the IAA is low. 
To address this challenge, we provide a \emph{semantically based methodology to identify 
%em concessions that presuppose argumentation 
argumentative concessions and compare them with other types of concessions} (Section \ref{section:typo}).  Second, the automatic retrieval of argumentative concessions through discourse markers is difficult due to the polysemy with contrast and other discourse relations \citep{prasad2014reflections}. State of the art discourse parsers trained on Penn Discourse Treebank \citep{Prasad08} achieve low accuracy in the identification of concessive uses of discourse connectives in general due to the small number of instances in the training data; this result is reflected also on our task of identifying \emph{argumentative concessions}. 
%em after revisions
Therefore, we first observe the distribution of concessions in manually labeled data both through expert annotation and through crowdsourcing (Section \ref{section:md}). We focus our analysis on four discourse connectives: \textit{but}, \textit{though}, \textit{however}, and \textit{while} for two reasons: (1) they constitute 85\% of the overall occurrences of potential markers of concessions in our dataset, and (2) they are highly polysemous and thus require disambiguation (Section \ref{section:data}). We used 20\% of the overall occurrences of these markers for our crowdsourcing study (i.e., 2440 instances) as well as a separate dataset of 1000 instances for the expert annotation task.  
%To address this challenge,
Second, as a step towards large-scale analysis of persuasive discourse, we use the manually labeled data as training, development and test sets to build computational models to detect argumentative concessions (Section \ref{section:empi}).  %We focus our analysis on four discourse connectives: \textit{but}, \textit{though}, \textit{however}, and \textit{while} for two reasons: (1) they constitute 85\% of the overall occurrences of potential markers of concessions in our dataset, and (2) they are highly polysemous and thus require disambiguation (Section \ref{section:data}). % EM_after_review While 
We achieve a moderate F1 of 57.4\% on the development set and  46.0\% on the test set via a self-training method (Section \ref{section:evaluation}).  
%our subsequent error analysis highlights that the self training method is able to generalize and identify other types of concessions that are argumentative, but were not considered in the annotation guidelines used to label our test set (Section \ref{section:evaluation}).
%SM the above is no longer true due to refine guidelines, no? 

Our findings from the manual labeling (using both expert and crowdsourcing annotations) indicate that the type of argumentative concessions we investigate is almost equally likely to be used in winning ($\Delta$-awarded) and losing (non-$\Delta$) arguments.
%SM here we need to also mention running automatic labeling when we get the results.
While this result seems to contradict theoretical assumptions,  we provide some reasons related to the nature of the ChangeMyView subreddit (Section \ref{section:discussion}). 
%SM I added this about ERIC
In addition, we present a preliminary analysis on running our computational models  on a different dataset, The Yahoo News Annotated Comments Corpus, \ ( \emph{ERIC}: Engaging, Respectful, and/or Informative Conversations \citep{napoles2017automatically}, where comments are labeled as persuasive or not via crowdsourcing (persuasion is a ``binary label indicating
whether a comment contains persuasive language or an intent to persuade''). Differently from delta points in ChangeMyView, the persuasiveness label does not inform us about what discourses achieved persuasion among the participants of an actual interaction, but provides hints as to what type of language is perceived as persuasive by third parties. We observe that argumentative concessions are present more in persuasive comments than in non-persuasive comments in ERIC dataset. 

The dataset and code are available at https://github.com/debanjanghosh/concessions.

%% file: related.tex
\section{Related Work} \label{section:related}

%EM: afterreview

Concessions have received various non-overlapping definitions. We take as a default definition the one provided by \cite{grote1997ma}, which focuses on the semantic relation holding between two connected propositions (A, B): 
``On the one hand, A holds, implying the expectation of C. On the other hand, B holds, which
implies not C, contrary to the expectation induced by A":

$ A \rightarrow C $    e.g., ``This dress is gorgeous" $\rightarrow$ it is worth buying  

$ B \rightarrow \neg C$ e.g.,  ``but, it is expensive" $\rightarrow$ it is NOT worth buying 

The conflict resides in the expectations generated by A and B which are mutually exclusive. Besides this concessive configuration, that is called indirect concessions \citep{azar1997concession,izutsu2008contrast}, there are direct concessions where \textit{B} and the following implication rule do not have to be verbalized. In these cases the main clause constitutes the negation of the expectation arising from proposition \textit{A} (e.g., ``This dress is gorgeous, but I am not buying it").
The distinction between direct and indirect concessions is particularly relevant from a discourse perspective since direct concessions are more suitable to be used nonargumentatively to describe states of affairs and to increase interest for an unexpected contradiction. 
%EM:  after review 
%have added brackets around B and the following implication rule since they do not
%always have to be verbalized. In a sentence such as ‘Although [he was tired]A, he went
%to the gym’, the main clause constitutes the negation of the expectation arising from
%proposition A, which would sound something like ‘if somebody is tired he does not want
%to go to the gym’. This concessive configuration has been called direct (Azar, 1997;
%Blakemore, 1989; Izutsu, 2008) since one of the connected propositions (A or B) ‘directly
%denies one part of the potential causal relation’ (Azar, 1997: 307), as opposed to indirect
%concessions where the conflict resides in the expectations generated by A and B which
%are mutually exclusive. As noticed by Izutsu (2008: 662), the assumption evoking for-
%mula does not hold with indirect concessions: the sentence ‘The car is stylish and spa-
%cious, but it is expensive’ does not generate the expectation ‘??If the car is stylish and
%spacious, (then normally) it is not expensive’ but two conflicting expectations (‘the car is
%worth buying’/‘the car is not worth buying’).

The pragmatic function played by concessions has been widely addressed in linguistically oriented \emph{Discourse Studies}. To cite just a few, drawing from Quintilian \emph{Institutio Oratoria}, \cite{perelman1971new} explains that through concessions, 
\begin{quote}
``one gives a favorable receipt to one's opponent's real or presumed argument. By restricting his claims, by giving up certain theses or arguments, a speaker can strengthen his position and make it easier to defend, while at the same time he exhibits his sense of fair play and his objectivity". 
\end{quote}

\cite{mann1988rhetorical} list concessions among presentational rhetorical relations, aimed at increasing the addressee's positive attitude towards the speaker's beliefs. 
According to Pragmadialecticians \citep{van2007argumentative} concessions are attested either in the confrontation stage of an argumentative discussion, where the difference of opinion between two parties is stated, and the opening stage, where the common starting points of the discussion are established. 
In a corpus study of  interview transcripts involving environmental activists,  \cite{uzelgun2015managing} show that the concessive construction ``yes \dots but" constitutes a privileged viewpoint to investigate the (dis)agreement space singling out what is accepted and what is criticized. \cite{antaki1999show} identify a series of rhetoric strategies through which concessions strenghten the speaker's positions undermining counterarguments. Similarly, \cite{couper2000concessive} exemplify how concessive repairs are used by speakers to back down their overstatements and foster credibility. 
%EM-after reviews: added following two references 
In their investigation of myside bias in written argumentation,  \cite{wolfe2009argumentation} show that texts which present and rebut other-side arguments achieve better ratings of agreement, quality and overall impression of the author. However, the presence of concessions preceding rebuttals did not lead to a higher perception of the arguments' quality. It has to be remarked that students that were asked to evaluate the texts were not individually engaged in an interactive conversation; therefore, face-threatening risks were not the same as those characterizing a face to face argumentative exchange. 
The argumentative value played by concessive discourse relations is recognized by \cite{green2010representation} who inserts concessions among the RST relations needed to represent argument presentation in a biomedical corpus. 
%EM following reference added

As far as datasets are concerned, while  resources annotated as to agreement and disagreement are provided  \citep{walker2012corpus}, cases of partial (dis)agreement are neglected. 
 Computationally, \cite{somasundaran2009recognizing} develop an unsupervised method for stance detection in online debates, taking into consideration also concessionary opinions. 
%EM 
From a pragmatic perspective, a growing interest is devoted to the identification of persuasive discourse strategies in order to implement classification experiments. \cite{young2011microtext} released a corpus of blog posts annotated as to persuasive tactics according to sociological studies (e.g., promises/threats, mentions to duties/generalizations, appeal to reason). The evaluations of the predictive power of the strategies for the identification of persuasion show that the label \emph{Reason} guarantees accuracy; however the unigram SVM baseline for \emph{Reason} is poor due to the difficulty in identifying rhetorical relations. Drawing from the same set of tactics \cite{young2011microtext} presented a corpus of 37 transcripts from four sets of hostage negotiation transcriptions annotated as to persuasion features. Using supervised learning algorithms they show that persuasion tactics constitute machine-learnable features. 
\cite{tanetal16} analyzed shallow linguistic and interactional features which happen to be persuasive in ChangeMyView, a subreddit where users exchange opinions and assign a \textit{Delta point} to the user that managed to change their view: dissimilarity at the lexical level seems to play a major role, together with the order and the number of interactions.
\cite{wei122016post} investigated the performance of different sets of features in predicting persuasion: they have found that argumentation-based features perform better than shallow textual features. \cite{habernal2016argument} have recently released a corpus of 16k pairs of arguments over 32 topics annotated as to persuasiveness using crowdsourcing. Annotators were also asked to provide reasons behind their choices. Experiments with feature-rich SVM and  Long Short-Term Memory (LSTM) neural networks \citep{hochreiter1997long} reveal that predicting persuasion is a task that requires analytic skills still hard to attain computationally.

The detection of rhetorical/argumentative relations
%EM-review: which seem to play a major role in persuasion,
is bound to the performance of state of the art discourse parsers trained on the Penn Discourse Treebank, the largest annotated corpus so far available (Prasad et al. 2009). \cite{pitler2009automatic} showed that syntactic feautures play a crucial role in disambiguating explicit connectives. Drawing from their work, \cite{lin2014pdtb} built a four-steps pipeline for the identification of implicit and explicit discourse relations passing through the identification of text spans functioning as argument. 
%EM: added 
Connectives indicating concessions have been proved by \cite{swanson2015argument} to significantly correlate with the presence of argumentation, even though not with argument quality. 
For the task of detecting the \emph{type} rather than the \emph{class} of discourse connectives, which is more similar in nature to our problem, the performance of state of the art models is still modest (56.91 F-measure even for explicit cases) \citep{biran2015pdtb}.  \cite{biran2015pdtb} treat PDTB discourse parsing as two separate tagging tasks and show that connective-specific grammatical features promise to improve the results.  On these grounds, we combine  various lexical patterns, pragmatic as  well as  semantic features in conjunction with a self-training approach for our task of identifying whether a discourse connective introduces an argumentative concession or not.

%% file: typology.tex
\section%{%A typology of concessions} 
{Argumentative concessions} \label{section:typo}
%EM afre review: I have embedded the definition of concessions in the previous paragraph
%Concessions are discourse relations that bind two propositions {\sc A} and {\sc B} so that {\sc A} implies the expectation of {\sc C}, while {\sc B}  states or implies {\sc not C} contrary to the expectation induced by A \citep{grote1997ma}. In a sentence such as ``[This dress is gorgeous]$_A$, but [it is expensive]$_B$" proposition {\sc A} entails the expectation ``The dress is worth buying", while proposition {\sc B} calls for the opposite expectation ``This dress is not worth buying". 
%EM-after review 
In order to answer our research question --- whether argumentative concessions increase persuasion --- we need to provide a definition of argumentative concessions. 
Drawing from \cite{musi2017did}, we consider concessions as displaying an argumentative function when the proposition introduced by the connective -- {\sc B} ---, which denies the expectations brought about by a preceding proposition, expresses the speaker's standpoint. This happens when the following two conditions are met:  (i) proposition {\sc B} is asserted by the speaker who is committed to its truth at the moment of utterance; and (ii) proposition {\sc B} is non factual --- its truth is not self-evident. Concessive relations such as ``Although it is already very warm, [there are no buds on the trees]$_B$" and ``He states that, despite the difficult situation,[the company will not go bankrupt]$_B$" are, therefore, not argumentative since propositions {\sc B} are an unassailable fact and a prediction with which the speaker does not necessarily agree, respectively. 
%EM-after review: add other type of concession
As far as proposition {\sc A} is concerned, what is conceded can either be: (i) a point that could \textit{possibly} be made by another speaker (i.e. ``[Sure, we could argue about some hypothetical other religion in the region causing similar problems]$_A$, but that hypothetical world is not this world'') or that belongs to \textit{common ground knowledge} (e.g., ``[He's no Hitler, of course]$_A$, but only because North Korea isn't powerful enough to start annexing its neighbors and has no substantial minority population to send to death camps''), or (ii) a claim previously made by another speaker in the discussion. Concessions of the type (i) can be used to prevent a counterargument. However, they do not necessarily display an interactional value, since the potential disagreement imagined by the speaker may have never happened. Concessions of the type (ii) are, instead, used in conversations as mitigating strategies to avoid disruptive disagreement. They, are, therefore, inherently argumentative: 

Example: 

Speaker 1:  ``[...] Basically, I'm glad Jackson opted to make excellent **movies,** instead of attempting a shot for shot visualization of a **book,** and I think the extended cuts subvert his success in that regard. [...] ''

Speaker 2: `` As far as Peter Jackson's opinion, the opinion of the author or producers of a content doesn't dictate how it should be interpreted. He is a person who thinks A, that doesn't mean that A is the be all and end all. Like any piece of media once it's released into the public it becomes its own beast that can have its interpretations judged by others.  [I agree with you that the quality of the movie matters]$_A$, but [if both versions are good movies and didn't make unnecessary changes then I think they are different rather than one being better]$_B$ ''.  

As explained by \cite{couper2000concessive} this concession type is meant to be persuasive:  a speaker, recognizing the validity of a point made by the hearer before expressing disagreement, avoids face-threatening acts and is perceived as reasonable by the hearer. We, thus, consider the description provided by 
Interactional Linguistics \citep{couper2000concessive,couper2005linguistic} as our definition of argumentative concessions: 

%Therefore proposition {\sc A} is presented in this type of concessions as given, but it  cannot be anaphorically retrieved in a preceding comment. In those cases the communicative intention of the speaker is that of anticipating a potential counterargument

%These conditions happen to be respected when concessions are used in conversations as mitigating strategies to avoid disruptive disagreement. According to Interactional Linguistics \citep{couper2000concessive,couper2005linguistic}, concessions are formed by three conversational moves: 

\begin{quote}
\begin{itemize}
\item 1st move: Speaker1 states something or makes some point 
\item 2nd move: Speaker2 acknowledges the validity of this statement or point (the conceding move)
\item 3rd move: Speaker2 goes on to claim the validity of a potentially contrasting statement or point.
\end{itemize}
\end{quote}

%As explained by \cite{couper2000concessive} this concession type is meant to be persuasive:  Speaker2, recognizing the validity of a point made by the hearer before expressing disagreement, avoids face-threatening acts and is perceived as reasonable by the hearer.
%Among argumentative concessions, two main types can be singled out according to the the semantic type of proposition {\sc A} and their information status. 
%In the first type proposition {\sc A} acknowledges the validity of a point made by another speaker. 
At a semantic level, proposition  {\sc A} (the conceding move of Speaker2) is always an evaluative proposition since the speaker positively qualifies a standpoint advanced in the preceding post directly through the expression of positive sentiment (e.g., ``I like your post, but [...]") or agreement (e.g. ``You are right, but [...]"). 
%EM-review: 
At an informational level, the sentiment of the evaluation constitutes new information in the discourse flow, while what is evaluated constitutes old information which coincides with a point that has been previously made by another speaker.   
%while the evaluation in itself constitutes new information, the target of the evaluation is given information since coincides with an anaphorically available point.
%SM what is propositon A; is it 2nd move of Speaker2 above? If soe we need to connect to that  

%In the second type proposition {\sc A} acknowledges the validity of a point that could potentially be made by a speaker or that belongs to common ground knowledge. Proposition {\sc A} can thus be both an evaluation (e.g., ``I am not an expert, but [...]") or, more generally, a shared interpretation (e.g. ``We all know that, but [...]") conveying information presented either as new or given. 

%Drawing from the hypothesis made in Interactional Linguistics we have decided to empirically test whether this type of argumentative concessions in a polylogical context, such as ChangeMyView, bear a persuasive force.  
%Due to the dialogical nature of the subreddit and the hypothesis made in Interactional Linguistics, we have decided to focus on argumentative concessions of the first type, (called from now on \textit{argumentative concessions}) in order to empirically test their persuasive force.

%% file: empirical.tex
\section{Data} \label{section:data}

To empirically test our theoretically-informed hypothesis we use ChangeMyView (CMV) subreddit ``dedicated to the civil discourse of opinions, and built around one simple idea: in order to resolve our differences, we must first understand them''.  Users start a discussion thread expressing their opinion (original post, OP) about a certain issue and other users challenge their view in a network of subsequent comments. ChangeMyView constitutes a particularly suitable environment for the study of persuasive argumentation. %for two reasons.
First, the users, in order to get their posts
published, have to respect a series of submission and comment rules, which ensures the presence of argumentation:\footnote{The following quotations are taken from the CMV wiki (https://www.reddit.com/r/changemyview/wiki/index).}
\begin{quote}
\begin{itemize}
\item Submission rules:
\begin{enumerate}
\item[(s1)] Try to explain the reasoning behind your view, not just what that view is (500+ characters required).
\item[(s2)] You must personally hold the view and be open to it changing.
\item[(s3)] Submission titles must adequately sum up your view and include CMV: at the beginning.
\item[(s4)]  Only post if you are willing to have a conversation with those who reply to you, and are available to start doing so within 3 hours of posting.
\end{enumerate}
\item Comment Rules: 
\begin{enumerate}
\item[(c1)] Direct responses to a CMV post must challenge at least one aspect of OP’s stated view (however minor), or ask a clarifying question.
\item[(c2)] Don't be rude or hostile to other users.
\item[(c3)] Refrain from accusing OP or anyone else of being unwilling to change their view.
\item[(c4)] If you have acknowledged/hinted that your view has changed in some way, please award a delta ($\Delta$).
\item[(c5)] Comments must contribute meaningfully to the conversation. 
\end{enumerate}
\end{itemize}
\end{quote}

This combination of submission and comment rules matches the main rules for conducting an ideal critical discussion \citep{van1996fundamentals}: rules (s1), (c2) and (c3) constitute an application of the freedom rule (``Parties must not prevent each other from advancing standpoints or from casting doubt on standpoints''), while rule (s4) guarantees that the burden of proof rule (``A party that advances a standpoint is obliged to defend it if asked by the other party to do so'') is respected. Rule (c1) can be interpreted as a sort of standpoint rule (``A party's attack on a standpoint must relate to the standpoint that has indeed been advanced by the other party''). Finally, rule (c4) presupposes the closure rule according to which ``A failed defense of a standpoint must result in the party that put forward the standpoint retracting it''. 
%(collected from both the Delta-awarded and no-Delta awarded posts).
%em after review 
In addition, ChangeMyView constitutes a unique environment for the study of persuasion: although corpora annotated as to the presence of persuasive language exist (e.g., \cite{napoles2017finding}), CMV, as far as we know, is the only available dataset containing first-hand information (through awarded $\Delta$ points) as to what arguments are perceived as persuasive by actual language users. 
%the awarded $\Delta$ points provide information about the arguments that are perceived as persuasive by the actual users: they are attributed by the author of the original post to the argument that has changed his view, who has to specify how and why his view has changed.

%EM -after review
Finally, since the discussed issues cover very different topics and the participants who are anonymous can have different epistemic backgrounds, the results of the analysis in terms of persuasion strategies can be generalized across different text genres and contexts. 
In our study, we used a dataset collected from the ChangeMyView platform introduced by \citet{tanetal16},  where only the replies by the root challenger are considered, defining all the replies by the root challenger in a path as the rooted path-unit. For each rooted path-unit that wins a $\Delta$, they select a rooted path-unit in the same discussion tree that did not win a $\Delta$ but was the most ``similar'' in topic (similarity computed using Jaccard similarity measure). With this setup, the goal is to de-emphasize what is being said, in favor of how it is expressed. We thus end up with a paired dataset that contains $\Delta$-awarded and no-$\Delta$ comments, which can be used to test the hypothesis that argumentative concessions are persuasive strategies by computing their distribution in the $\Delta$-awarded and no-$\Delta$ comments.  

We focus our analysis on four discourse markers: \textit{but}, \textit{though}, \textit{however}, and \textit{while} for two reasons: (1) they constitute 85\% of the overall occurrences of potential markers of concessions in our dataset (see Table \ref{table:markers}), and (2) they are highly polysemous and thus require disambiguation. For each marker, we  collect the sentence in which the marker is occurring as well as the previous and the next sentence (from both $\Delta$ and no-$\Delta$ comments). 

\begin{table}[h!]
\begin{center}
\begin{tabular}{|c|r|r|}
\hline \bf Marker & \bf $\Delta$ & \bf no-$\Delta$ \\ \hline
admit & 26 & 17 \\
albeit & 9 & 17 \\
although & 78 & 93\\
but & 4403 & 5908 \\
concede & 8 & 13 \\
despite & 89 & 114\\
even if & 255 & 314 \\
even though & 101 & 129 \\
even when & 31 & 55 \\
however & 132 & 213 \\
in spite of & 10 & 8 \\
nevertheless & 3 & 10\\
notwithstanding & 1 & 4\\
non the less & 0 & 0\\
nonetheless & 7 & 18 \\
the fact remains that & 3 & 4\\
though & 426 & 619 \\
whereas & 48 & 73\\
while & 575& 763\\
\hline
\end{tabular}
\end{center}
\caption{Distribution of candidate concessions markers in ChangeMyView}
\label{table:markers}
\end{table}

Next, we manually annotated two datasets using expert annotation and crowdsourcing, respectively. The annotators were asked to identify whether a marker introduces an argumentative concession (\emph{arg\_c}) or not. The first dataset includes a set of 1,000 examples of the 4 discourse markers and has been annotated by an expert annotator.% as to whether the markers introduce argumentative concessions (\emph{arg\_c}) or not.
	The second dataset consists of two samples each containing 10\% of the overall occurrences of the four discourse markers from $\Delta$ and no-$\Delta$ comments, resulted in a total of 1,220 instances each (total of 2440 instances). One of the sample will be used as development set ($dev$) and one as test set ($test$) in the machine learning experiments. The $dev$ set will be used for tuning the parameters of the learning model, while the $test$ set will constitute the blind set used to evaluate the learning model. Instead of expert annotators, we use a \emph{crowdsourcing platform} -- Amazon Mechanical Turk (MTurk) to identify the \emph{arg\_c}. The task is framed as follows: Given a sentence or pairs of two adjacent sentences in ChangeMyView containing one of the four discourse markers, five Turkers on (MTurk) are asked to identify and label as \emph{arg\_c} those occurrences in which the sentence preceding the connective expresses agreement or positive sentiment towards a point previously made by another speaker.  The Turkers are provided with detailed instructions of the task and multiple examples.
%EM after review
Before conducting the full crowdsourcing annotations tasks, we ran a pilot annotation to evaluate whether Turkers are able to complete such task. In the pilot annotation round, we realized that providing the Turkers with the mere definition of argumentative concessions was not enough since annotators were selecting as argumentative concessions also those conceding statements from fictive speakers or belonging to common knowledge. Therefore, we refined the guidelines to discard this latter type of concessions pointing to the fact that they tend to contain modal adverbs of certainty (e.g., \textit{(for) sure, of course}) that, differently from markers that imply a dialogical exchange (e.g., \textit{agree}, \textit{understand}), signal that the truth of a general statement is taken for granted since it belongs to the common ground (e.g., ``of course, runners use the lower body more than the upper body, but the same is true for swimmers in reverse (they use more upper than lower)''). We, moreover, specified that they contain second person pronouns/adjectives that do not work as deictics, but are used impersonally to portray fictive scenarios (i.e., ``yes, a portion of \textit{your} money goes to things you don't support; but likewise, things \textit{you} do support are partially funded by other people who may not support those things.'').  For both $dev$ and $test$ sets, containing 1,220 sentences (or pair of sentences) each, we obtain 6,100 labels from the Turkers. To assure a level of quality control, only qualified Turkers were allowed to perform the task (i.e., more than 95\% approval rate and at least 5,000 approved Human Intellingence Tasks --- HITS). Each HIT contained one sentence (or pair of sentences) to be labeled and the Turkers were paid 5 cents for each HIT.
%EM_afte_review: this is true both for dev and for test data? 
We compute the inter-annotator agreement (IAA) between the Turkers via Fleiss Kappa measure \citep{fleiss1971measuring}.
%EM_after_review
We obtain fair agreement both for $dev$ ($\kappa$ = 0.31) and for $test$ ($\kappa$ = 0.22) sets. 
%EM_after_review 
As underlined by \cite{passonneau2014benefits}, standard measures for inter-annotator reliability are not suitable to account for corpus quality, and in addition, the $\kappa$ values obtained on semantic annotation tasks is not high.
Therefore, we consider the obtained IAA as a reasonable output which does not undermine the relevance of the corpus. 

\section{Distribution of argumentative concessions in manually labeled data} \label{section:md}

The expert annotated dataset contains of 229 instances of \emph{arg\_c} and 751 instances of other types of concessions or instances expressing contrast or other discourse relations, which we denote as \emph{other}. 
For the crowdsourcing experiment, we take majority voting among the five Turkers to choose the label  (\emph{arg\_c} or \emph{other}). We obtained 201 \emph{arg\_c} instances (out of 1,220) for the $dev$ set and 174 \emph{arg\_c} instances (out of 1,220) for the $test$ set. 
%development data and 174 \emph{arg\_c} instances (out of 1,220) for test data.
To better understand the difficulty of the task, we selected the cases where 3 annotators chose one label, while 2 others choose the other label and we asked an expert annotator to label these cases. We compared the expert labels with the labels obtained by majority voting (i.e., the labeled chosen by 3 Turkers). 
% EM_after/review 176 \emph{arg\_c} instances (out of 1,220).
%As a further validation, we asked an expert annotator to provide a meta-annotation %EM: added the overall number as required
%for the cases where only three Turkers agreed on the same label: 
%EM_after review 17 
In the $test$ set, out of 225 such instances, in 22 cases there is a mismatch between the label annotated by three Turkers and the expert annotator; in the $test$ set, out of 280 such instances, the mismatch amounts to 67 cases. Zooming into the mismatched examples, in the $dev$ set 4/22 cases are not recognized as argumentative concessions by the Turkers, while 18 cases are misclassified as argumentative concessions. In the $test$ set, the type of mismatch seems more balanced: 31/67 instances are not recognized as argumentative concessions by Turkers, while 36/67 are misclassified as argumentative concessions. We observe that discarding these most confusing cases where majority is formed by only 3 Turkers, the IAA improves both in the $dev$ set ($\kappa$ = 0.43) and in the $test$ set ($\kappa$ = 0.35).
%SM if the number in Table 6 are after expert annotatord correction add this sentence below and fix the numbers above : see my previous comment
Taking into account the annotation provided by the expert for those instances, the number of \emph{arg\_c} in $dev$ and $test$ sets is 179 and 168, respectively.  
%26 out of the 201 instances annotated as argumentative concessions by Turkers were not, while 3
%EM after_review 18 
%examples annotated as \emph{other} were argumentative concessions. 

From the qualitative analysis, it seems that argumentative concessions are misclassified when the proposition functioning as standpoint (proposition B)  expresses a negative evaluation of the opinion held by another speaker, while the  proposition A is used to specify the degree of such an attack, more than to express a partial agreement. In a couple of sentences such as ``I don't believe the the war was worth the suffering, but to say that nothing positive happened because of it is a little disingenuous'', the speaker's intention is not that of partially agreeing with the judgment made by the preceding speaker, but to clarify to what extent it is indeed ingenuous. Cases not recognized as argumentative concessions contain a positive sentiment rather than an explicit agreement with what was said by the preceding speaker (e.g., ``You have noble goals, but there are very real downsides''). 

The distribution of argumentative concessions in threads that have and have not been awarded a $\Delta$ point does not appear to be significantly skewed amounting to 99 (awarded) vs 130  (not awarded) occurrences in the set annotated by an expert and 165 vs., 194 in the crowdsourcing experiment.
Due to the limited quantity of annotated data available, before discussing the relevance of the attested distribution (section \ref{section:discussion}), we develop a preliminary computational model to classify and trace back argumentative concessions on a larger scale. 

%The final number of \emph{arg\_c} instances amounts to 
%EM after_review 
%178. 

%EM: what about the development set? 
%Interestingly, the cases of mismatch with the expert annotator were those where only 3 Turkers agreed. 

%maybe put a discussion here 

%Fleiss $\kappa$ is 0.3 showing fair agreement \cite{fleiss1971measuring}.

\section{Computational models to detect argumentative concessions} \label{section:empi}

%questions on numbers - 1235 
% is the 10% of all.
%our computational models.

We seek to automatically identify argumentative concessions from the CMV corpus. We frame the task as a binary classification task: \emph{arg\_c} vs. \emph{other}.
Since using expert annotators is an expensive process, 
we use the expert annotated data as a small labeled training data (total of 980 instances; Section \ref{section:md}\footnote{After annotation we noticed that 20 examples were duplicates, and thus we removed them from the final set}). Due to this small annotated data scenario, we develop a self-training method \citep{clark2003bootstrapping,mihalcea2004co}, which uses the remaining unannotated 70\% of the CMV corpus (section \ref{section:automatic}) as unannoted data. From our annotation studies it is clear that all the datasets are highly unbalanced (the size of the \emph{arg\_c} class is much smaller than the size of the \emph{other} class).
%we only annotate around 10\% of the CMV corpus as training data. 
%The dev and test sets (10\% of the CMV corpus each) are annotated via crowdsourcing (Section \ref{section:data}).

The data annotated through crowdsourcing (the two samples of 10\% of the CMV corpus each) is used as development ($dev$) and test ($test$) sets, respectively.
  %and ran various experiments. \footnote {Recall, for both development and for test we have annotated 10\% of the CMV data (i.e., a total of 20\% of CMV data).} 

In the following section we discuss the features as well as the linguistic patterns associated with the argumentative concessions used in our experiments (section \ref{section:description}). Second, we discuss the classification task using self training in section \ref{section:automatic} and the results in section \ref{section:evaluation}, including comparison with an off-the-shelf state-of-the art discourse parser trained on Penn Discourse Treebank that aims to classify the \emph{type} of discourse connectives, not their \emph{class} \citep{biran2015pdtb}.

\subsection{Feature Description} \label{section:description}
  
We use linguistically-motivated features inspired by research on discourse relation identification and persuasive argument identification \citep{stab2014annotating,ghosh2016coarse}. A brief description of the features is given below.
%SM I used discoure relation identification, can you add citations to that? Biran work, Ani Nenkova etc

%are motivated by recent research on discourse and persuasive argument  identification \citep{stab2014annotating,stabgurevych2016,ghosh2016coarse}. Short descriptions of the features are below.

\begin{itemize}
%SM I replace $all$ with entire since it was not clear what "all" means. 
\item \textit{bag-of-words:} We selected the entire CMV dataset used by \cite{tanetal16} to extract the bag-of-words features (e.g., unigrams and bigrams). We consider every sentence that contains the four candidate discourse markers (i.e., ``but'', ``while'', ``however'', and ``though'') as candidate examples. If the sentence starts with the markers we also consider the previous sentence. Next, based on tf-idf scores (i.e., we treated each sentence as a document), we select the top 1,000 unigrams and bigrams as candidate features. %Words with highest tf-idf scores are ``agree'', ``think'', ``could'' etc.      
\item \textit{personal pronouns and adjectives:} By definition, argumentative concessions dialogically point to the stance taken by the previous speaker. They, therefore, contain personal pronouns and adjectives (i.e., ``I see your point'' etc.): we consider as features both a list of first person (i.e., ``I'',``me'', ``my'',``mine'') and second person (i.e., ``you'', ``your'', ``you' re'') pronouns and adjectives. 

%It is common to use pronouns in argumentative concessions (i.e., ``you'' are right; ``I'' see ``your'' point etc.) so we have a list of first person (i.e., ``I'',``me'', ``my'',``mine'') and second person (i.e, ``you'', ``your'', ``youre'') pronoun lists as features.
\item \textit{modal verbs:} modal verbs (e.g., ``could'', ``should'') work as indicators of claims since they indicate that what is expressed in a proposition is not unassailable, but might be otherwise \citep{palau2009argumentation}. They, thus, frequently appear in propositions B of argumentative concessions which constitute the speakers' standpoints. We define a Boolean feature which indicates if a candidate example contains a modal verb.

%EM Modal verbs such as  ``could'' ``should'' are frequently used in argumentative concession to modulate the degree of certainty when expressing a claim 
 
%EM I checked and in the 2014 paper they did not use features still 

\item \textit{hedges:} Hedges are linguistic devices used to mitigate the speaker's commitment to the truth of a proposition \citep{hyland1996talking}, i.e., ``I \textit{tend to} accept''. They include possibility modals next to other linguistic items expressing the degree of speaker's certainty. They can be ambiguous depending on constructional features: the propositional attitude indicator \emph{I think} works, for instance, as a hedge in parenthetical constructions (e.g., ``It is not worth buying it, \emph{I think}''), while it merely signals subjectivity when used as a main verb (e.g., ``I think it is not worth buying it''). 
The use of hedges is common in argumentative concessions since they contribute to avoid a potentially face-threatening act of abrupt disagreement.  
%indicate a uncertainty about the argument --- 
%``I \textit{tend to} accept'' -- which is common in concessions. 
\cite{tanetal16} argue that depending on the context, hedges can make an argument weaker or easier to accept by softening its tone. 
%\cite{tanetal16} argues that hedges often  make an argument easier to accept by softening its tone. 
Based on their research and also on \cite{hanauer2012hedging}, we collect a set of candidate hedge cues and use them as Boolean features (presence or absence of a hedge word).  
\item \textit{Jaccard Similarity:}
In argumentative concessions proposition {\sc A} always  expresses positive sentiment towards a claim expressed by the previous speaker. 
%Argumentative concessions often repeat the previous claim and then presents an alternate.
Thus, we use Jaccard Similarity to measure lexical similarity between the sentence/sentences containing the candidate concession and sentences from the original post (we removed stopwords). We use the \textit{maximum} similarity value as a feature. 
%how similar is the concession to the main post. 
%For each instance, we measure similarity between each sentence of the previous comment (e.g., the comment to which \emph{arg\_c} is part of a reply post) after removing stopwords and then use the \textit{maximum} similarity as a feature. 
\item \textit{sentiment feature:}
By definition, argumentative posts tend to contain opinion on the other posts. A post that is tagged with subjectivity will be a useful feature to identify concessions. We use (a) the MPQA Lexicon \citep{wilson2005recognizing} of over 8,000 positive, negative, and neutral sentiment words, (b) an opinion lexicon with around 6,800 positive and negative sentiment words \citep{hu2004} to see whether training instances contain sentiment words. 
%EM: should we justify the use of sentiment words? 
%SM I am not sure what we did here" just see if the example contains sentiment words? 
%\item{Lexical Template:}
\end{itemize}

Apart from the above features, we also retrieved lexical patterns that could be indicators of argumentative concessive uses of the discourse markers. These patterns are used in proposition {\sc A} to express a positive evaluation about another speaker's claim. 
% EMthat may represent argumentative concessions. 
%These patterns are used as \emph{high precision} markers
They could assist in achieving higher precision in identifying \emph{arg\_c} instances. Below is a short description of the semi-automatic retrieval of lexical patterns.       

\paragraph{Semi-automatic retrieval of lexical patterns} 

We present a bootstrapping algorithm that automatically learns lexical patterns expressing argumentative concessions.  The algorithm begins with only two seed phrases -- ``I agree'' and ``you are right''-- the most common two patterns expressing argumentative concessions according to annotation results done by an expert on a separate development set of one-hundred utterances (not used in the above annotation studies).
We used 80\% of our ChangeMyView data (i.e., all except the development as well as the test dataset) for the bootstrapping algorithm. The algorithm operates on a simple structural assumption: In argumentative concessions, words forming the proposition positioned before the marker (for ``but'', and ``though'') or in the marker's scope (for ``while'') express agreement with a point made in a preceding comment. To identify such patterns, first, we retrieve these words (e.g., all words in the proposition before ``but'').  
Second, we extract all possible trigrams, four-grams, and five-grams from this set of words and search for the two seed phrases in the ngrams. For instance, we identify all instances of ``[$\dots$] I [$\dots$] agree [$\dots$]'', where [$\dots$] represent zero or more occurrences of words. We obtain patterns such as ``I agree [completely]'' or ``[I think] you are right [about]''. Third, using each new pattern we attempt to identify new lexical constructions. For example, given the pattern ``I agree [completely]'' we search for the occurrences of ``I [$\dots$] completely'' where [$\dots$] represent zero or more occurrences of any words, excluding negation. With this search, we arrive at new patterns, such as ``I [understand] completely'' and ``[I think] you are [correct]''.

Since this method relies on structural syntactic similarity, we embed the following semantic rules: (i) keep only patterns, which contain propositional attitude indicators (i.e.,  verbs ``think'', ``realize'', and the constructions ``be right'', ``be correct'')  or indicators of sentiment (i.e., verbs ``love'', ``like'') and (ii) select the patterns which contain the pronoun ``you'' or the adjective ``your''-- to retain just those seeds where the target of agreement is an opinion held by another poster. 
Finally, in the fourth stage, using the new lexicon we again resume the search for new patterns (i.e, using ``[$\dots$] I [$\dots$] realize [$\dots$]'') and this process continues until we do not find any new patterns. As a result of the bootstrapping algorithm we obtain 329 lexical patterns, which we call $B\_Lexicon$. We, then, manually filter the bootstrapped lexicon to eliminate redundancies, merging patterns which instantiate the same linguistic constructions and adjustments based on the dev set (e.g., removing double negatives "I don't disagree"). As a result of manual filtering we finally obtained 116 lexical patterns indicating argumentative concessions ($B\_Lexicon_{MF}$). Table \ref{table:lexicons} shows some examples from this manually filtered lexicon. 

\begin{table} [t]
\centering
\begin{scriptsize}
\begin{tabular}{||p{3.5cm}||} 
\hline
Manually filtered lexicon\\
\hline
I would agree with you \\
I fully agree that \\
I see what you \\
I see where you \\
I think you are correct \\
\hline
\end{tabular}
\end{scriptsize}
\caption{Examples of manually filtered lexicon}
\label{table:lexicons}
\end{table}

%SM it is very confusing the different splits of that data. It will be helpful to have a table in the Data Section to mention all info there. 

\subsection{Self-training method for identifying concessions}
\label{section:automatic}

%Our task is a binary classification task: label an instance as \emph{arg\_c} or \emph{other}. 
%SM I mentioned this before

%The class weights are inversely proportional to the number of instances in the categories.  We employ $\chi^2$ based feature selection and used ``top 300'' features for all the experiments. %As stated earlier, we use the 70\% remaining data (i.e., apart from the training and test data) for classification via self-training. 

Self-training algorithms \citep{mihalcea2004co,clark2003bootstrapping} start with a small subset of annotated training data and attempt to increase the amount of training data by using a large set of unannotated data. We adopt the approach of \cite{mihalcea2004co} that used self-training as ``a tagger that is retrained on its own labeled cache on each round''  \citep{mihalcea2004co,clark2003bootstrapping}.  Similar to \cite{mihalcea2004co}, we start with a small set of labeled data that is the training data  ($L$), in our case the data labeled by the expert annotator consisting of 980 instances. We build a classifier using the linguistically-motivated features described above, and then apply the learned model on a set of unlabeled data ($U$). For our experiments, the $U$ is the 70\% of the CMV data (i.e., apart from the training, test, and dev sets; each is 10\% of the CMV data). Now, instead of classifying directly on the total set of data $U$, we split $U$ in $P$ random pools where each  pool contains $U'$ unlabeled instances. From each $P$ pool we select only those $G\_c$ argumentative concession instances and $G\_nc$ non-argumentative concession instances with a labeling confidence exceeding a particular threshold. Similar to \cite{mihalcea2004co}, while adding the new instances to the training data $L$, we maintain the original class distribution of training data between \emph{arg\_c} and \emph{other} categories. The threshold could be a preset value of probability; in this experiment, we varied the number of data instances to select while keeping the original class distribution. These instances (i.e, $G\_c$ and $G\_nc$) in turn are added to the original training set $L$. The classifier is retrained with the new set of training data (i.e, $L$ + $G\_c$ + $G\_nc$) and this process continues for all the $P$ pools. Note, in each step we also evaluate the classifier on the dev set to assess the quality of the new data that is added to the training set $L$ (Table \ref{table:selfresults}). For details of the self-training procedure, please see \cite{mihalcea2004co,clark2003bootstrapping}.   
We use the Support Vector Machines classifier with RBF kernel (we use Scikit-learn tool \citep{scikit-learn}). The class weights are inversely proportional to the number of instances in the categories. %As stated earlier, we use the 70\% remaining data (i.e., apart from the training and test data) for classification via self-training. 
%SM I do not understand what is G_c as 10 or 50 in table? From text it means is the instance added to training set, but then it does not add up. if you add 10 to 220 shoild be 239... So something is not explained well above. 

As a final classifier we used a system combination: if any dev/test instance contains a lexical pattern obtained via bootstrapping described in the previous section, we classify that instance as \emph{arg\_c}, otherwise,  use the decision of the self-training classifier. %Since we have gold annotation (i.e., training data) of only 980 instances we employ a self-training procedure using the CMV data. 
The dev data is used to fine tune all parameters in our experiment and to choose the \emph{best} parameters (see next section for detailed discussion). % Table \ref{table:selfresults} show that the best parameters are . % to conduct experiment on the test data. The self-training process is described next.  

%\paragraph{Parameters in the self-training process}

%We set two different parameters in this experiment. Pool size ($P$) is the number of examples selected from the unlabeled set $U$ for annotation at each iteration.
% Growth size (G) Number of most confidently labeled examples that are added at each iteration to the set of
%labeled data L.
%As previously noticed (Ng and Cardie, 2003), there
%is no principled method for selecting optimal values for these parameters, which is an important disadvantage of these algorithms.
\section{Experiments and Results} \label{section:evaluation}

\begin{table*} [t]
\centering
%\begin{small}
\begin{tabular}{||c|c|c|c|c|c|c||}
\hline
\multicolumn {2}{||c|}{Self-training Setting} & \multicolumn{2}{c|}{Optimal size (training)} &  \multicolumn{3}{c||}{Performance (Max. F1)} \\
Pool size & \emph{$G\_c$} & \emph{arg\_c} & \emph{other} & P & R & F1  \\
\hline
50 & 10 & 430 & 952 &  56.9 &  57.5 &  \textbf{57.2}  \\
100 & 10 & 441 & 821 &  56.7 &  57.0 &   56.8  \\
1000 & 10 & 289 & 811&  65.9   &   45.8   &  54.0   \\
2000 & 10 &  229 & 751 & 65.3 &     44.6 &     53.0  \\
100 & 50 & 442 & 964 & 64.5 &   51.3 &    \textbf{57.4} \\
500 & 50 &   424 & 946 & 63.9 &   48.0 &   54.8 \\
1000 & 50 &  415 & 937 & 64.1     & 47.5     & 54.5 \\
2000 & 50 &  279 & 801 & 63.1     & 46.3    &  53.4 \\
\hline
\end{tabular}
\caption{Experimental results of the Self-training method on the dev set ({\bf bold} are best scores)}
\label{table:selfresults}
\end{table*}

Before reporting the results of our self-training classifier and our baselines, we report the results of an off-the-shelf parser that aims to label the \emph{types} of discourse connectives \citep{biran2015pdtb}, noted as $Ob_{parser}$ in Table \ref{table:baseline}.
%baseline efforts (Table \ref{table:baseline}). 
%We first use the discourse parser \citep{biran2015pdtb} to identify the concessions (i.e, for brevity, $Ob_{parser}$ in Table \ref{table:baseline}).
%EM_after-review We have to explain that this is not a parser  that can be considered as a baseline. 
\cite{biran2015pdtb} have trained the parser on the Penn Discourse Treebank (PDTB) corpus \citep{miltsakaki2004penn} using discourse connective features, lexical features, syntactic features etc. Their model can identify the \emph{types} of discourse relations, such as \textit{Concession}, \textit{Contrast}, \textit{Pragmatic Concession}, and \textit{Pragmatic Contrast} that \emph{types} of the \emph{class} of discourse relation \textit{Comparison}. The off-the-shelf parser results in a low F1 measure of only 6 with precision of 13.2 and recall of 4 for the dev data.\footnote{Since the performance of the off-shelf-parser on the development data is very low compared to the other methods we did not conduct any further experiment on the test set.} This low performance is not unexpected. First, the parser is trained on the PDTB corpus, which is based on of Wall Street Journal (WSJ) articles and the language is vastly different from the ChangeMyView subreddit. In addition, PDTB has a small number of  concessions in general and it is not clear how many of those are argumentative concessions if any. 
%EM-after review: Debanjan, is it up to sixteen right? 
%Second, this parser is trained to classify up to sixteen level 2 subtypes of PDTB relations instead of binary classification task (i.e., concession vs. no-concession). Given the number of training data on concessions are few in PDTB it is no surprise that when the parser is tested on a different corpus the performance is poor. 
%EM_after_review (Maybe)? 
It has to be noted that state of the art discourse parsers do not take into account different pragmatic values underlying a discourse relation. The level of granularity required to classify different types of concessions is higher than that necessary to classify what discourse relation is conveyed by a single connective. Nevertheless, the performance achieved by our system is comparable to that achieved by \cite{biran2015pdtb} in the classification of explicit discourse relations (56.91 F1). 

\paragraph{Baselines.} We use two baselines: 1) a rule-based system based on the lexical patterns learned via bootstrapping ($B\_Lexicon_{MF}$ in Table \ref{table:baseline}), and 2) the system combination that uses the $B\_Lexicon_{MF}$ and a SVM classifier that uses all the features but without the self-training process ($SVM_{noST}$). %Finally, the third row presents the performance of the rule-based system that just uses the lexical patterns (i.e, $B\_Lexicon_{MF}$). 
The performance of both of these baselines is better on the $dev$ set than on the $test$ set, which is expected as $dev$ set was used to check the final pattern lexicon. The recall of the  $B\_Lexicon_{MF}$ patterns is low for the test data, meaning that there are different patterns not covered in our collection of lexical patterns and argumentative concessions can exist in many ways that are not retrieved by simple lexical rules. %The performance of the system combination SVM classifier is better on the dev set than on the test set. Since we have tuned our parameters on the development data it was expected. 

%SM when looking at Table 3 you mention "training data". but what I asked is results of using lexical patterns as rule based (basically the first component of system combination).

\paragraph{Self-training.} Table \ref{table:selfresults} show the results of the system combination (lexical patterns and SVM classifier \emph{using self-training}) on the $dev$ set, used to decide the best parameters (pool size and the number of instances to add to the training set).
%SM is this just self training or also the system combination where first we test if the instance contains lexical pattern label as arg_c
In column one (i.e., \textit{pool size}) we report the number of random unlabeled instances (i.e, \textit{U'}) that were tested via the classifier. Column two represents the maximum number of instances ($G\_c$) that are added to the training data. 
For Recall, we also add $G\_nc$ to maintain same class distribution between \emph{arg\_c} and \emph{other} categories. The next two columns show the size of training data from our two classes respectively that achieve the highest F1 scores for the dev set. We start the experiments with the labeled training data by expert annotators (also used in the baselines) and gradually add $G\_c$ and $G\_nc$ that are predicted by the classifier. Finally, the last three columns show the P/R/F1 on the dev set. We report all the columns based on the highest F1 achieved by the models. For instance, in the first row of Table \ref{table:selfresults}, the pool size is 50 and $G\_c$ is 10. The size of the unlabeled data is close to 8,400 (after removing duplicates), which means here the number of pools is 168. Now, we classify each pool from the set of $P$, and depending upon the classification result, we add $G\_c$ (here, maximum is 10 instances) and $G\_c$ accordingly from each pool. After the classifier has evaluated a certain number of pools the F1 of 57.2 is achieved. Meanwhile, starting from 229 $arg\_c$, the size of the $arg\_c$ is now 430. We also observe a common trend between the various runs of the self-training procedure; the accuracy increases till a certain number of $G\_c$ + $G\_nc$ is added to the original training set of $L$, but after that the performance drops probably due to added noise.% that are different from the original training set.    
 The best model based on the dev set, has the pool size of 100 and $G\_c$ of 50 reaching F1  of  57.4 that is close to a 5\% improvement compared to the baseline results without self-training on the dev set. 

We used this best setting (i.e., pool size of 100 and $G\_c$ of 50) to run our system combination using self-training on the test set. We also used the dev data for feature selection. We employ $\chi^2$ based feature selection and observe that the ``top 300'' features that performed best for the dev data. Subsequently this setting (i.e, ``top 300'' features) is used for the test set. This feature set of "top 300" features contains modal verbs such as ``may'', ``could''; Jaccard similarity value; pronouns such as ``I'', ``your'; hedges such as ``almost'', ``probably'', ``somewhat''; words such as ``greatest'', ``excitement'', ``interesting'' from the sentiment lexicons; unigrams such as, ``agree'', ``recognize'', ``argument'', ``think'', and finally, bigrams such as, ``while you'', ``argument is'', ``i absolutely'', to name a few. The results of self-training show improvement over the baselines, but to a lesser degree when compared to $SVM_{noST}$ (2\%), and a higher degree when compared to $B\_Lexicon_{MF}$ (13.1\%).  

\begin{table*} [t]
\centering
%\begin{small}
\begin{tabular}{||c|c|c|c|c|c|c|c||}
%\begin{tabular}{||c|c|c|c|c|c||}

\hline
\multirow{2}{*}{Computational Model} & {Training size} & \multicolumn{3}{c}{$dev$} & \multicolumn{3}{c||}{$test$} \\
& (\emph{arg\_c};\emph{other}) & P & R & F1 & P & R & F1\\
\hline
$Ob_{parser}$ &  \_  & 13.2 & 4.0 & 6.0 & - & - & - \\

$SVM_{noST}$ &  (229;751) & 65.3 & 44.6 & 52.9 & 35.1 & 58.9 & 44.0\\
$B\_Lexicon_{MF}$  & \_ & 65.5 & 43.9 & 52.5 & 48.3 & 25.0 & 32.9\\
Self-training (best) & - & 64.5 &   51.3 &    \textbf{57.4} & 38.0 &     58.3 &   46.0 \\
\hline
\end{tabular}
\caption{Experimental results of classifiers on the $dev$ and $test$ set. The self-training results report the \textbf{best} performing parameters.} 
\label{table:baseline}
\end{table*}

%The results show that although the precision is  high, the recall is moderate. 
%SM I commented the sentence about precision since is not high

% results need to be updated 
To understand the quantitative results better,  % why the automatic classification of argumentative concessions still constitutes a difficult task 
we have randomly selected and qualitatively analyzed a sample of 135 classified occurrences from the development data. We compared recurrent characteristics both of occurrences that the system classified \emph{arg\_c} in accord with gold data and those that it classified as \emph{arg\_c}, while the human annotators did not. 
It turns out that occurrences classified as \emph{arg\_c}  both by the system and by the annotators tend to include lexical patterns, which unambiguously express agreement paired with second person pronouns/adjectives which anaphorically point to a previous claim
(i.e. ``I do \textit{agree} with a lot of points in \textit{your} post, but there is a huge disconnect between the title of your post'').
%EM_after_review 
Looking at occurrences that have been classified by the expert annotator as \emph{arg\_c} while considered by the system as \emph{other}, they tend to lack an unambiguous expression of agreement and an anaphoric reference to a preceding post. More specifically, agreement is expressed through modal adverbs expressing different degrees of certainty (i.e. ``\textit{Possibly}, but not because they're missing out on experiences''; ``\textit{Of course} it is but if you're okay with the humane treatment of animals to include population control  why are you uncomfortable with the humane treatment of animals in other contexts as the end result is the same?''). However, the state of affairs over which the modal adverbs have scope is elliptical, since shared as given information by the participants to the discussions (speakers as well as readers). In classification experiments ellipsis is a phenomenon hardly detectable since dependent on the pragmatic content of the utterance. 

%% file: discussion.tex
\section{Discussion: are concessions persuasive strategies?} \label{section:discussion}
%EM-review 
%As underlined in section \ref{section:related}, in theoretical studies concessions are considered as persuasive strategies. 
%since they embody the ``principle of charity" according to which making maximum sense of the thoughts of others we optimize agreement \cite{davidson2001inquiries}.
To test the theoretically-informed hypothesis that argumentative concessions work as persuasive strategies, we look at their distribution in the $\Delta$ awarded vs. no-$\Delta$ awarded comments. %To test if this hypothesis is empirically consistent we have  first looked at the distribution of argumentative concessions in
%EM after review 
First, we consider the manually labeled data. The small set of training data annotated by the expert annotator, as well as the dev and test sets annotated in the crowdsourcing experiment. The distribution of argumentative concessions in the $\Delta$ awarded vs. no-$\Delta$ awarded comments on all these datasets is given in Table \ref{table:DnoD}. Second, we look at the distribution when considering the  argumentative concessions \emph{predicted} by our best classifier (system combination with self training). We look at the dev and test set, as well as the rest of the unlabeled CMV corpus (Table  \ref{table:DnoD2}). As a caveat, we have to point out that the due to the low accuracy of classifier, the observed distribution based on the classifier predictions cannot be considered reliable. For example, the predictions of the computational model on the dev and test set miss to identify more than 1/3 of argumentative concessions compared to the gold annotations.    
%The distribution of the connectives that were labeled as argumentative concessions \emph{arg\_c} in the manually annotated data is presented in Table \ref{table:DnoD}, while that on argumentative concessions predicted by the system in Table  \ref{table:DnoD2}.
%training as well as in the test data is presented in Table \ref{table:DnoD}. 

\begin{table}[h]
\begin{center}
\begin{tabular}{ |c|c|c|c|c|c|c|}
\hline \bf Marker & \bf $\Delta$-training & \bf no-$\Delta$-training & \bf $\Delta$-dev & \bf no-$\Delta$-dev & \bf $\Delta$-test & \bf no-$\Delta$-test\\ \hline
but & 39 & 59 & 68 & 83 & 83 & 82\\
however & 25 & 3 & 3 & 5 & - & -\\
though & 22 & 27  & 4 & 8 & 2 & 1 \\
while & 13 & 13 & 2 & 5 & - & - \\
\hline
total & 99 & 130 & 78 & 101 & 85 & 83\\
\hline
\end{tabular}
\end{center}
\caption{Distribution of argumentative concessions in $\Delta$ and no-$\Delta$ comments in the manual labeled datasets }
\label{table:DnoD}
\end{table}

\begin{table}[h]
\begin{center}
\begin{tabular}{ |c|c|c|c|c|c|c|}
\hline \bf Marker & \bf $\Delta$-dev & \bf no-$\Delta$-dev & \bf $\Delta$-test & \bf no-$\Delta$-test & \bf $\Delta$-unlabel & \bf no-$\Delta$-unlabel\\ \hline
but & 49 & 57 & 51 & 47 & 1100 & 793\\
however & 2 & 4 & - & - & - & -\\
though & 4 & 5  & - & 1 & 8 & 2 \\
while & 1 & 2 & - & - & - & - \\
\hline
total & 56 & 68 & 52 & 48 & 1112 & 795\\
\hline
\end{tabular}
\end{center}
\caption{Distribution of \emph{predicted} argumentative concessions in $\Delta$ and no-$\Delta$ comments  }
\label{table:DnoD2}
\end{table}

%While in test data the argumentative concessions seems to be slightly more frequent in the $\Delta$ than in the no-$\Delta$, that is not the case for the training data. 
Overall the numbers suggest a fairly equal distribution of argumentative concessions in the $\Delta$ and no-$\Delta$ comments both in the manually labeled data and in those predicted by the classifier. We have tested the statistical significance of the distribution of argumentative concessions on the manually labeled sets using $\chi^2$ test: the results are not significant at p $<$ 0.05 on training and test sets, while they are significant on dev set, where argumentative concessions are less frequent in winning arguments. 

%Both in the training and the test data argumentative concessions are are slightly more frequent in negative than in positive threads. 
These results suggest that argumentative concessions do not increase persuasion in ChangeMyView, challenging the assumptions made in the rhetorical literature.
However, they do not allow us to draw conclusions scalable to different contexts about the persuasive role played by this type of concessions. 
%However,  does not allow to draw conclusions about the non persuasive role played by this type of concessions regardless of the context.
They rather provide a further confirmation that the persuasive value played by lexical meta-discursive features is context-bounded and crucially depends on the the rhetorical situation.  For example, hedges seem to increase persuasiveness in scientific writing (\cite{hyland1996writing}, while they decrease it in other kinds of messages (\cite{blankenship2005role}. 
% However, these two manually annotated dataset are small. To be able to validate it on the larger ChangeMyView data, we needed to observe the distribution on a larger sample. Therefore, we apply our classification method to the entire ChangeMyView dataset. The results show that argumentative concessions are almost equally as likely to appear in a winning ($\Delta$) and losing thread (no-$\Delta$): \textit{but}($\Delta$) = 671/5888 (11\%), no-$\Delta$ = 531/4392 (12\%); \textit{though} $\Delta$ = 98/619 (16\%), no-$\Delta$ = 78/496 (16\%); \textit{while} $\Delta$ = 31/575 (5\%), no-$\Delta$ = 43/762 (6\%). 
% It has to be remarked that our method achieves only moderate accuracy. However, the results are consistent with the manually annotated data, thus providing a further confirmation that argumentative concession are not preferably associated to threads which obtained a $\Delta$. 
% This means that argumentative concessions do not increase persuasion in ChangeMyView.
%challenge the assumptions made in the rhetorical literature, but it does not allow to draw conclusions about the non persuasive role played by this type of concessions regardless of the context.
%EM after_review 
%\footnote{Since our computational models achieve only moderate performance and since the entire unannotated dataset from ChangeMyView was also using during self-training we felt it was not meaningful to report the distribution results on the entire ChangeMyView using the computational models.}
When it comes to argumentative concessions, they embody what in the philosophical tradition has been called \textit{principle of charity} or \textit{principle of rational accommodation} according to which ``we make maximum sense of the words and thoughts of others when we interpret in a way that optimizes agreement'' \citep{davidson2001inquiries}: conceding the claims made by the author of the original post the speaker shows his intention of interpreting them in the best possible way. In doing so he reinforces his ethos, presenting himself as a reasonable discussant. At the same time, he minimizes the risk of a face-threatening opposition which could arise from the apparent incompatibility with the statement expressed in the nucleus (proposition B). The principle of charity has, in fact, to be understood as a methodological presumption that guarantees the understanding of another point of view in its argumentatively strongest form to allow a possibly adequate critique and reach agreement through persuasion. 
	This principle is structural in ChangeMyView, being at the core of the subreddit mission of resolving differences of opinions starting from a deep understanding of them. Users who choose to write on this subreddit have to respect the submission rules and are, thus, by default charitable. Therefore, linguistic strategies such as argumentative concessions that encode the principle of charity constitute persuasive strategies in the speakers' mind, but are plausibly perceived as routinized expressions by the addressees.  In other words, they do not shape the argumentative profile of single users, but are conventional strategies belonging to the activity type envisioned by ChangeMyView.

As stated in the introduction, the lack of corpora with explicit signals as to what pieces of discourse achieved persuasion makes it difficult to investigate the persuasive roles played by concessions in datasets belonging to different discourse genres and dialogue activity types. To our knowledge, beside the ChangeMyView subreddit, the only other corpus labeled as to persuasive features is  the Yahoo News Annotated Comments Corpus (YNACC)  \citep{napoles2017automatically}).
This corpus contains around 140K threads, taken from the comments sections of Yahoo  News  articles as well threads from the Internet Argument Corpus. 
Discussion posts are annotated by expert and untrained (i.e., Turkers) annotators with specific labels, 
both at the thread (e.g., constructiveness) and at the comment level (e.g., topic). Among the latter type, persuasiveness is defined as a ``A binary label indicating whether a comment contains persuasive language or an intent to persuade.'' Differently from \emph{Delta points} in ChangeMyView, the persuasiveness label does not inform us about what discourses achieved persuasion among the participants of an actual interaction, but provides hints as to what type of language is perceived as persuasive by third parties. In other words, we experiment with this dataset to assess whether argumentative concessions are perceived by language users as persuasive strategies, namely strategies used by speakers as an attempt to persuade, despite the actual pragmatic outcome. 
%persuasive, agreement/disagreement, controversial, etc.
We selected a subset of YNACC %for our experiments 
that are annotated by expert annotators. This subset contains 4,719 posts that are annotated with the label ``persuasive'' and 17,616 posts that are annotated with the label ``not persuasive''. From this subset we evaluate only the posts that contain the selected markers for our experiments, ``but'', ``while'', ``though'', and ``however''. 

We use the same training data (described in Section  \ref{section:empi}) to predict the binary label \emph{arg\_c} and \emph{other} from the YNACC posts. We use the same features that are described in Section \ref{section:description} except the \emph{Jaccard Similarity} feature since the YNACC dataset does not indicate which post is a reply to another post. After removing the duplicates from the YNACC posts we observe that out of 649 persuasive posts, 321 (49.4\%) are classified as \emph{arg\_c} (i.e., concessions) whereas out of 1,263 non persuasive posts, 417 (33\%) posts are classified as concessions.
According to these results, argumentative concessions are deemed as persuasive strategies, since they correlate with the persuasiveness label.
In Table \ref{table:ericexplicit} we present the count of the main %occurrence of the main explicit 
expressions of concessions across the persuasive and the non persuasive datasets: ``?'' depicts one or zero occurrence of the particular pattern. For example, ``pattern\_acknowledge'' shows that any occurrence of the expression, ``I'' and following either of ``concede'', ``acknowledge'', and ``think'' is regarded as a concession of the pattern ``pattern\_acknowledge''. Note in the above case, ``also'' or ``too'' can appear between the above two words.
%Zooming into the different types of linguistic constructions playing the role of argumentative concessions, there seems to be difference in how proposition A is realized:
%em : plan here is to insert a table with the frequency of constructions used in our argumentative lexicon. 
\begin{table}[h]
\begin{center}
%\begin{tabular}{ ||c|c|c|c||}
\begin{tabular}{||p{3cm}|p{5cm}|p{2cm}|p{2.2cm}||}
\hline
Explicit\_expressions (patterns) 	&	Description	&	\multicolumn{2}{c||}{Count} \\
\cline{3-4}
& & persuasive & not persuasive \\
\hline
%persuasive (Count)	&	not persuasive (Count)	\\
pattern\_yes	& [yes$|$sure$|$of course$|  $correct$|$right$|$true] [,]	&	24.81	&	13.62 \\			
\hline
pattern\_acknowledge	&	[I] ?[also$|$too] [concede$|$acknowledge$|$think]	&	12.63	&	9.58	\\
\hline			
pattern\_see	&	I ?[adverb$|$modal] [see$|$get] [?] [you$|$your]	&	8.47	&	4.51	\\
\hline							
%pattern\_correct	&	you ? [correct$|$right]	&	3.85	&	1.82	\\						pattern\_respect	&	I [respect$|$understand$|$appreciate] [?] [that$|$what$|$why$|$where] [you$|$your]	&	2.47	&	1.19\\
pattern\_agree & I ?[adverb$|$modal] [agree] & 0.77 & 0.63 \\
\hline							
\end{tabular}
\end{center}
\caption{Distribution (in \%) of explicit expressions (patterns) of concessions in persuasive and not persuasive instances in ERIC corpus}
\label{table:ericexplicit}
\end{table} 
%EM: I thik pattern_yes and pattern_correct should most likely go together and I think we should better specify pattern respect (right now "where" is not clear ---it is where you are going right?). In terms of proportions there seems not be much difference among the patterns but we can safely say concessions are considered as persuasive strategies. 
Other patterns of concessions also exist (e.g., ``I appreciate your'') but the frequency is very low. Overall, it seems that regardless the type of linguistic construction at work, concessions are are more frequent in persuasive threads. 